\PassOptionsToPackage{table}{xcolor}
\documentclass[10pt,twocolumn,letterpaper]{article}
\usepackage{iccv}
\usepackage{color}
\usepackage{times}
\usepackage{epsfig}
\usepackage{graphicx}
\usepackage{amssymb}
\usepackage{amsmath,bm}

\usepackage{booktabs,multirow}
\usepackage{algorithm}
\usepackage{algorithmicx}
\usepackage{algpseudocode}
\usepackage{graphics}
\usepackage{threeparttable}
\usepackage[normalem]{ulem}
\usepackage{float}
\usepackage{amsfonts}
\usepackage{subfig}
\usepackage{enumitem}
\usepackage{array}
\usepackage{colortbl}

\definecolor{bblue}{rgb}{0,0.588,0.90}
\definecolor{mygray}{gray}{.9}
\makeatletter
\newcommand{\thickhline}{%
    \noalign {\ifnum 0=`}\fi \hrule height 1pt
    \futurelet \reserved@a \@xhline
}
\def\ourdataset{HIDE}
\makeatother
\newcommand{\tabincell}[2]{\begin{tabular}{@{}#1@{}}#2\end{tabular}}

\usepackage[pagebackref=true,breaklinks=true,letterpaper=true,colorlinks,bookmarks=false]{hyperref}

 \iccvfinalcopy % *** Uncomment this line for the final submission

\def\iccvPaperID{2850} % *** Enter the ICCV Paper ID here

\ificcvfinal\fi
\begin{document}

%%%%%%%%% TITLE
\title{Human-Aware Motion Deblurring}
\author{Ziyi Shen$^{1,2}\thanks{The first two authors contributed equally to this work.}$~\hspace{1pt}, Wenguan Wang$^{1,2*}$,  Xiankai Lu$^{1}$, Jianbing Shen$^{1,2}$\thanks{Corresponding author: \textit{Jianbing Shen}.}~, Haibin Ling$^{3}$,  Tingfa Xu$^{2}$, Ling Shao$^{1}$\hspace{1pt}   \\
	\small{$^1$} \small Inception Institute of Artificial Intelligence, UAE  \hspace{0pt}
	\small{$^2$} \small Beijing Institute of Technology, China \hspace{0pt} \small{$^3$} \small Stony Brook University, USA \\
 {\tt\small \url{https://github.com/joanshen0508/HA_deblur}}}

\maketitle
\thispagestyle{empty}

%%%%%%%%% ABSTRACT
\begin{abstract}
	This paper proposes a human-aware deblurring model that disentangles the motion blur between foreground (FG) humans and background (BG).
	The proposed model is based on a triple-branch encoder-decoder architecture. The first two branches are learned for sharpening FG humans and BG details, respectively; while the third one produces global, harmonious results by
	comprehensively fusing multi-scale deblurring information from the two domains.
	The proposed model is further endowed with a supervised, human-aware attention mechanism in an end-to-end fashion. It learns a soft mask that encodes FG human information and explicitly drives the FG/BG decoder-branches to focus on their specific domains.
	%
	%Above designs lead to a fully differentiable motion deblurring network, which can be trained end-to-end.
	%
	To further benefit the research towards Human-aware Image Deblurring, we introduce a large-scale dataset, named \ourdataset, which consists of 8,422 blurry and sharp image pairs with 65,784 densely annotated FG human bounding boxes.
	\ourdataset~is specifically built to span a broad range of scenes, human object sizes, motion patterns, and background complexities.
	Extensive experiments on public benchmarks and our dataset demonstrate that our model performs favorably against the state-of-the-art motion deblurring methods, especially in capturing semantic details.
	\vspace{-4mm}
\end{abstract}

%%%%%%%%% BODY TEXT
\vspace{-0pt}
\section{Introduction}
\vspace{-0pt}
Image deblurring, \ie recovering a sharp latent image with significant details from a single degraded image, has long been an active research area in computer vision.
With the increasing use of handheld devices, such as cell-phones and onboard cameras, motion blur has become a ubiquitous problem to confront with.
%
%The blur may be caused by atmospheric turbulence (in astronomy), defocusing, as well as the relative motion between the camera and the scene.
In this work, we focus on the dynamic scene deblurring problem and propose a human-aware deblurring model by \textit{explicitly discriminating between the blurred FG humans and BG}.
Basically, due to the relative motion between a camera and objects, FG and BG often undergo different types of image degradation~\cite{pan2016soft,nah2017deep}.
In addition, according to the imaging mechanism, each independent object experiences a varied motion blur as well, due to the specific distance between the object and image plane.
Among different objects, human beings are the most common and essential in our daily lives.
%
%Humans carry rich and strong semantics and structures (such as human motion patterns, clothing textures, \etc), which are informative for recovering details.
Humans are often accompanied by
unpredictable perturbations, thus providing a representative example for the in-depth study of the dynamic deblurring problem. In addition, restoring humans in a scene has broad
application prospects in tasks such as pedestrian detection.
Furthermore, with the dramatic increase in popularity of live shows and hand-held devices,
massive amounts of human-focused photos and short videos
have been created (since humans in these setting often draw viewer attention~\cite{bylinskii2016should}). A specially designed and human-aware deblurring approach would be highly beneficial for processing and editing human-focused visual data.

\begin{figure}[t]
	\centering
	\footnotesize
	\includegraphics[width=1\linewidth]{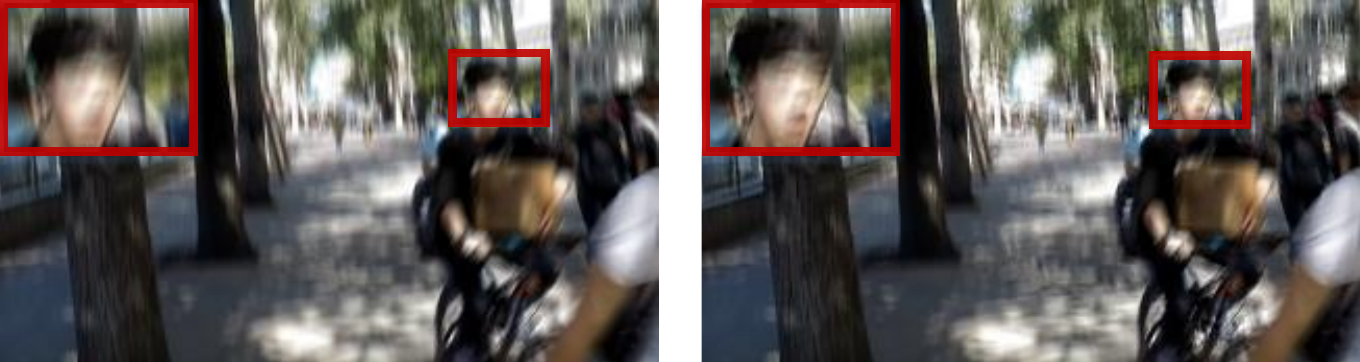}
	\vspace{0.3pt}
	\mbox{}\hfill (a) Blurred image \hfill\mbox{}
	\mbox{}\hfill (b) Sun \etal~\cite{sun2015learning}\hfill\mbox{}
	
	\includegraphics[width=1\linewidth]{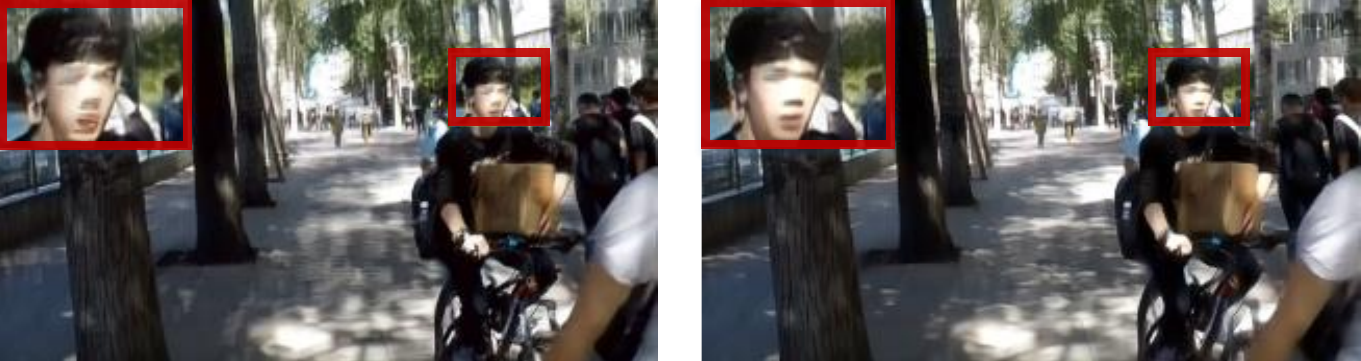}
	\vspace{0.3pt}
	\mbox{}\hfill (c) Nah \etal~\cite{nah2017deep}\hfill\mbox{}
	\mbox{}\hfill (d) Kupyn \etal~\cite{Kupyn18}\hfill\mbox{}		
	\includegraphics[width=1\linewidth]{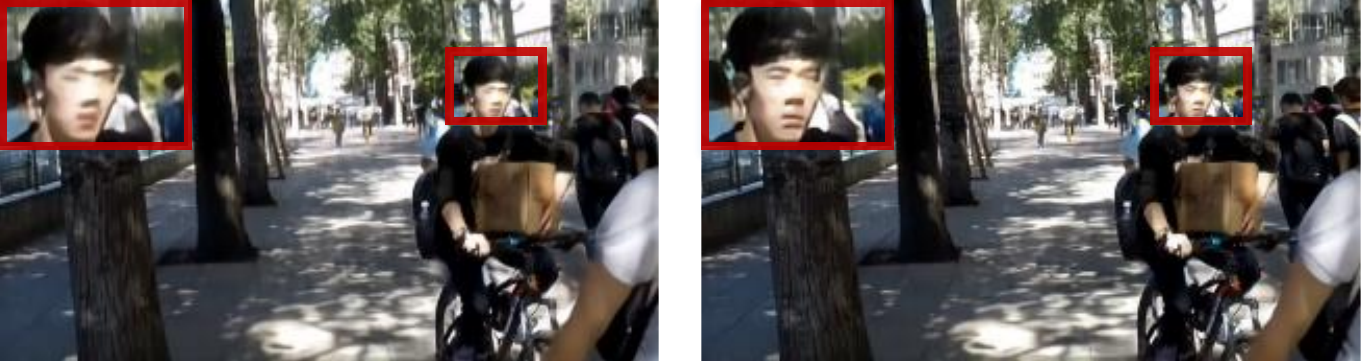}
	\vspace{-2pt}
	\mbox{}\hfill (e) Tao \etal~\cite{Tao18} \hfill\mbox{}
	\mbox{}\hfill (f) Ours \hfill\mbox{}	
	\vspace{-6pt}
	\caption{\small A challenging blurred image undergoes heterogeneous blur caused by camera motion and human movement.}
	\vspace{-10pt}
	\label{fig:top}
\end{figure}

Most existing non-uniform deblurring models~\cite{nah2017deep} attempt to deblur the FG and BG simultaneously. However, this leads to inferior performance and potential artifacts due to the neglection of the multiple motion patterns.
Only a few pioneering heuristic methods~\cite{pan2016soft,hyun2014segmentation} estimate object motion blur kernels.
However they do not emphasize the importance and particularity of human-focused deblurring and instead rely solely on pre-computed FG masks (\eg, Fig.~\ref{fig:top}).

Though their promising results do address the value of handling FG/BG blurs separately, the last generation of deblurring models put FG/BG information aside in favor of directly learning a uniform blur kernel using neural networks.
We believe that the main reasons for this radical choice are the lack of \textbf{(1)} an effective method for incorporating FG/BG information into neural networks in an end-to-end manner, and \textbf{(2)} a large-scale, carefully designed deblurring dataset with FB/BG annotations.

To tackle the first issue, we propose a novel human-aware attention guided deblurring network that
learns and leverages FG human and BG masks to explicitly capture the heterogeneous FG/BG blurs in a fully differentiable and end-to-end trainable manner.
More specifically, our model is built upon a fully convolutional encoder-decoder scheme.
It is equipped with a \textit{differentiable and supervised attention mechanism}, which is specially designed for learning a soft human-mask and can be trained end-to-end.
Based on this attention network design, we further extend our model with a \textit{multi-head decorder} structure containing three branches. The first two decoder branches are used to explicitly model the FG human and BG blurs, respectively, and the last one is designed for collecting and fusing both FG and BG multi-scale deblurring information and producing a final harmonious deblurring result for the whole scene.
Here, the human-aware attention acts as a gate mechanism that filters out unrelated encoder features and allows the FG/BG decoder-branches to focus on their specific domains. By comprehensively fusing the deblurring features from different domains, it is able to reconstruct the image with explicit structure and semantic details. Such a design leads to a unified, human-aware, and attentive deblurring network.  By explicitly and separately modeling the human-related and BG blurs, our method can better capture the diverse motion patterns and rich semantics of humans, leading to better deblurring results for both FG humans and BGs.

\begin{table*}[t]
	\centering
	\resizebox{0.99\textwidth}{!}{
		\setlength\tabcolsep{2.5pt}
		\renewcommand\arraystretch{1.0}
		\begin{tabular}{r||c|c|c|c|c|c|c|c|c}
			\hline\thickhline
\rowcolor{mygray}
			Dataset~~~~~~~~ & Pub. & Year & \# Images &Resolution &Systhesis &Motion Description&Content &\tabincell{c}{Pub. \\Ava.}&\tabincell{c}{Fore. \\Anno.} \\
\hline
\hline
		  BM4CS~\cite{kohler2012}&ECCV&2012&$4\!\!\times\!\! 12$ &$800\!\!\times\!\!800$ &Convolution  &Camera Motion: 6D Trajectories &Natural Images&\checkmark&\\
			
			VOC-Sampled~\cite{sun2015learning}&CVPR&2015&1,000 & $\sim\!\!500\!\!\times\!\!300$ &Convolution &Camera Motion: Rotation \(\&\) Translation &Static Object  \(\&\) Scenes &\checkmark&\\
			
			BSD-Sampled~\cite{gong2017motion}&CVPR&2016&$200\!\!\times\!\! 10k$ &$300\!\!\times\!\!460$ &Convolution  &Camera Motion: Rotation \(\&\) Translation &Static Object \(\&\) Scenes&\checkmark&\\
			%\hline
			GoPro~\cite{nah2017deep}&CVPR&2017&3,214&$1280\!\!\times\!\!720$ &Integration &Dynamic Scenes& Outdoor Scenes&\checkmark&\\
MSCNN(WILD)~\cite{noroozi2017motion}&GCPR&2017&-&$1280\!\!\times\!\!720$ &Integration &Dynamic Scenes&Outdoor Scenes&&\\
			\textbf{HIDE (ours)} &-&2019&8,422&$1280\!\!\times\!\!720$ &Integration &Human Motion \(\&\) Dynamic Scenes&Pedestrians in Outdoor&\checkmark&\checkmark\\
			\hline
		\end{tabular}
	}
	\vspace{-8pt}
	\caption{\small Summary of existing popular non-uniform deblurring datasets and our proposed \ourdataset~dataset (see \S\ref{sec:relatedwork}).}
	\vspace{-10pt}
	\label{table:blurdataset}	
\end{table*}

To address the second issue, we introduce a large-scale dataset, \ourdataset, which is specifically designed for \textit{Human-aware Image DEblurring}. The dataset contains 8,422 pairs of realistic blurry images and the corresponding ground truth sharp images, which are obtained using a high-speed camera. Each pair of images is further combined with densely and professionally annotated FG human bounding boxes.
Additionally, these image pairs are intentionally collected to cover a wide range of daily scenes, diverse FG human motions, %poses,
sizes, and complex BG. %Our dataset will be made publicly available.
The components described above represent a complete image deblurring dataset, which is expected to advance this field.

In summary, our \textbf{contributions} are four-fold:
\begin{itemize}[leftmargin=*]\vspace{-0.09in}
\item A human-aware attentive deblurring network is proposed to explore the task of motion deblurring by explicitly disentangling the blurs of FG humans and BG.\vspace{-0.1in}
\item A differentiable, supervised attention mechanism is integrated for the first time to enable the network to exclusively concentrate on FG human and BG areas.\vspace{-0.1in}
\item A novel multi-head decoder architecture is explored to explicitly model FG/BG motion blur and comprehensively fuse different-domain information for global and harmonious deblurring.\vspace{-0.1in}
\item A large-scale dataset, \ourdataset, is carefully constructed for human-aware image deblurring, covering a wide range of scenes, motions, \etc, with densely annotated FG human bounding boxes.%\vspace{-0.05in}
\end{itemize}	

%-------------------------------------------------------------------------
\vspace{-3pt}
\section{Related Work}\label{sec:relatedwork}
\vspace{-3pt}
This section first provides a review of previous representative image deblurring datasets as shown in Table~\ref{table:blurdataset}, followed
by a survey of recent image deblurring models and a brief overview of differentiable neural attention.

\noindent\textbf{Image Deblurring Datasets:}~Image deblurring has experienced remarkable progress in recent years. One of the critical factors bootstrapping this progress is the availability of large-scale datasets.
Several early works~\cite{kohler2012,sun2015learning} directly convolved sharp images with a set of pre-defined motion kernels to synthesize blurry images. For instance, the BM4CS dataset~\cite{kohler2012} contains 48 blurred images generated by convolving four natural images with twelve 6D trajectories (representing real camera motions) in a patch-wise manner.
Similarly, Sun \etal~\cite{sun2015learning} built a larger dataset, which has 1,000 images sourced from the PASCAL VOC 2010 dataset~\cite{VOC2010}. Though widely used, such patch-wise generated datasets yield discrete approximations of real blurry images with pixel-wise heterogeneous blurs. Later, Gong \etal~\cite{gong2017motion} used 200 sharp images and 10,000 pixel-wise motion maps to develop a new dataset, by associating each pixel with the corresponding motion vector. Recently, to construct a more real blurry image dataset, several researchers~\cite{nah2017deep,noroozi2017motion} have generated dynamic blurred images by averaging multiple successive frames captured by high frame-rate video cameras. More specifically, the GoPro dataset~\cite{nah2017deep} contains 2,103 pairs of blurred and sharp images in 720p quality, taken from 240 fps videos with a GoPro camera. A similar strategy was adopted in building the MSCNN (WILD) dataset~\cite{noroozi2017motion}.

Despite having greatly promoted the advancement of this field, current datasets seldom target the task of human-aware deblurring with ground truth FG annotations. This severely restrains the research progress towards a more comprehensive understanding of the underlying mechanisms of motion blur. This work proposes a new dataset, \ourdataset, which is carefully constructed for human-aware deblurring and expected to inspire further explorations of non-uniform motion blur caused by object motion.	
%. It has 8,422 blurry and sharp image pairs with densely annotated FG human bounding boxes and fine-grained body keypoints, and is expected to inspire further explorations of non-uniform motion blur caused by object motions.	

\noindent\textbf{Blind-Image Deblurring:}~For the uniform blur problem, conventional methods typically resort to natural image priors to estimate latent images~\cite{FergusSHRF06,KrishnanTF11,ShanJA08,Michaeli-ECCV-2014,XuZJ13,ChoL09,Sun2013,XuJ10,ren2018deep}.
Furthermore, rather than simply assuming the whole image is associated with a uniform blur kernel, some methods estimate a global blur kernel descriptor~\cite{whyte,gupta}, or predict a set of blur kernels~\cite{hirsch,harmeling}. However, they are limited by assuming that the sources of blurs are camera motions.
For more general dynamic scenes with object motion blending, some other methods~\cite{hyun2013dynamic,shi2014,pan2016soft,ren2017partial} estimate patch-wise blur kernels, assuming different positions with the corresponding uniform blurs. They deblur the background and object regions separately by relying on pre-generated segments~\cite{pan2016soft,hyun2013dynamic}, or estimating motion flow to facilitate blur kernel prediction in a segmentation-free framework~\cite{hyun2014segmentation}.
More recently, with the renaissance of neural networks, several researchers~\cite{sun2015learning,gong2017motion} have turned to using deep learning to predict patch-wise blur kernels. Such methodology hinges on an intermediate for the final reconstruction. Numerous CNN-based methods have also been applied in an end-to-end fashion for image processing and generation problems, such as segmentation~\cite{lu2019see,wang19learning},  super-resolution~\cite{xu2019sr,lai2017fast,kim2016accurate,kim2016deeply,wang2015deep}, denoising~\cite{mao2016image,xu2017multi,zhang2017beyond,xu2018external,xu2018trilateral}, dehazing and deraining~\cite{ren2018gated,yang2017deep}, enhancement~\cite{ren2019low,yang2018image} \etc. In a similar spirit, more advanced deep learning based deblurring models~\cite{shen2018deep,xu17learning,nah2017deep,Kupyn18,Zhang18,Tao18} have been designed, \eg, the coarse-to-fine framework~\cite{nah2017deep}, recurrent neural networks~\cite{Tao18,Zhang18}, and adversarial learning~\cite{Kupyn18}.

The promising results of these CNN-based models demonstrate well the benefits of exploring neural networks in this problem. However, in general, they do not take into account different FG human motion patterns or BGs, nor address human-aware deblurring.
A few heuristic studies~\cite{hyun2013dynamic,shi2014,pan2016soft} have addressed the use of FG information.
These methods are effective on the images with a wide-range scene or undergoing a straightforward defocus blur.
However, the diacritical mechanism relies heavily on the method of segmentation and fails to learn a robust solution for multi-motion superposition in real dynamic scenes.

%While the diacritical mechanism heavily relies on the segmentation fashions which fail to learn a robust solution on a multiple motion superposition in the real dynamic scene.
%
Furthermore, their significant feature engineering, high computational cost, complicated pipeline, and dependency on segmentation pre-processing, limit the performance and applicability od these methods.
In this work, in addition to contributing a human-aware motion deblurring dataset, we explore FG/BG information by integrating a trainable attention mechanism into a fully convolutional encoder-decoder architecture.
By associating the encoder with soft attention-based FG and BG masks, a multi-head decoder is developed to enable explicit FG and BG blur modeling, and improve the performance, simultaneously.

\noindent\textbf{Differentiable Attention in Neural Networks:}~Recently, differentiable neural attentions have gained significant research interest.
They mimic the human cognitive attention mechanism, which selectively focuses on the most visually informative parts of a scene.
They were first explored in neural machine translation~\cite{bahdanau2014neural}, and later proved effective in various natural language processing and computer vision tasks, such as image captioning~\cite{xu2015show}, question answering~\cite{yang2016stacked}, scene recognition~\cite{cao2015look,wangresidual}, fashion analysis~\cite{wang2018attentive}, \etc.
In the above studies, an attention mechanism is learned in a goal-driven, end-to-end manner, allowing the network to concentrate on the most task-relevant parts of the inputs.

We propose an essential attention mechanism, called human-aware attention, which explicitly encodes FG human information by learning a soft human-mask.
It drives our FG/BG decoders to focus on their specific domains and suppresses irrelevant information.
Instead of learning attention implicitly, as done in the above mentioned methods, our attention module is learned from human annotations in a \textit{supervised} manner.
Additionally, the attention mechanism keeps our model fully differentiable, enabling it to be trained end-to-end.
As far as we know, this is the first time an attention mechanism is leveraged for image deblurring.	

\vspace{-2pt}
\section{Proposed \ourdataset~Dataset}\label{sec:dataset}
\vspace{-3pt}
%As a dynamic blur is caused by the relative movement between an imaging device and a scene, we formulate the blur generative model, in terms of the depth, as
%\vspace{-3pt}
%\begin{equation}\small
%{\delta_c}={\rm{F}} \cdot ({{\Delta x}}/{L}),
%\label{eq:blur}
%\vspace{-3pt}
%\end{equation}
%where ${\rm{F}}$ denotes the focal length, ${L}$ is the depth, ${\delta _c}$ and ${\Delta x}$ denote the blur and actual offsets in the scene, respectively.
\begin{figure*}[t]
	\centering
	\footnotesize
	\includegraphics[width=0.99\textwidth]{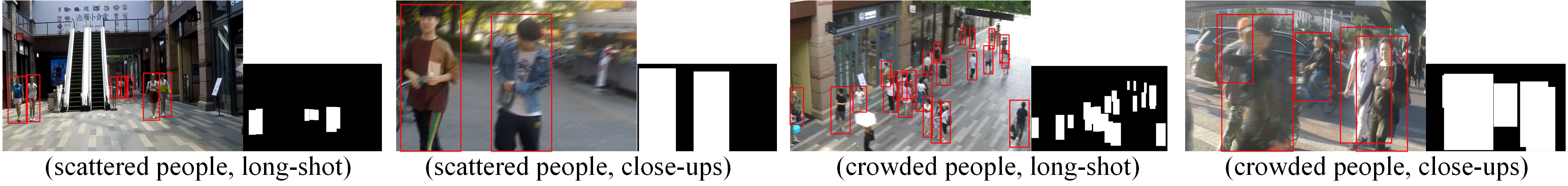}
\vspace*{-9pt}
	\caption{\small Example images from our \ourdataset~dataset with human bounding box annotations, human masks and attributes (see \S\ref{sec:dataset}). }
\vspace*{-14pt}
	\label{fig:sample}
\end{figure*}
%\ziyi{In real dynamic scene, images often undergo a series of blurs ${\Delta x}$ that consist of camera shaken ${\Delta x_a}$ and active object movement ${\Delta x_b}$.
%	%	
%	For images covering a wild-range scene, it is often modeled on the basis of the assertion that camera disturbance~\cite{whyte} dominates the cause of blur.
%	%
%	Such datasets~\cite{lai2016com,Sun2013} aims to simulate this simplified camera-driven condition.
%	%
%	To feature a more realistic situation, GoPro dataset~\cite{nah2017deep} further proposes to display the dynamic scene involving extra moving objects.
%	%
%	We note that images, especially for the close-up ones, usually undergo large motion blurs ($\delta_c$ is increased by the decline of factor $L$), which is caused by both camera motions as well as initial object movements.
%	%
%	However, the GoPro dataset~\cite{nah2017deep} mainly contains wide-range scene images but ignores close-range scene.
%	%
%	Therefore, it fails to fully reflect the dynamic blurs caused by the passive device interference and the initiative actions.
%}
Dynamic blurs are caused by the relative movement between an imaging device and a scene, mainly including camera shaking and object movements. Most representative datasets~\cite{lai2016com,Sun2013} were constructed on the basis of a simplified camera-driven assumption that camera disturbance dominates the cause of blur~\cite{whyte}. To model a more realistic situation, the GoPro dataset~\cite{nah2017deep} further proposed to display the dynamic scenes with extra active actions. However, it is mainly concerned with wide-range scenes, ignoring significant FG moving objects, especially in close-up shots. To fully capture the dynamic blurs caused by the passive device interference and initiative actions, our \ourdataset~dataset is elaborately collected to cover both wide-range and close-range scenes and address human-aware motion deblurring.
%%
%We note that images, especially for the close-up ones, usually undergo large motion blurs ($\delta_c$ is increased by the decline of factor $L$), which is caused by both camera motions as well as initial object movements.
%%
%
%%
%Therefore, it fails to fully reflect the dynamic blurs caused by the passive device interference and the initiative actions.
%}
%%
%To extract moving objects in these realistic scene, that undergo multiple blurs for dynamic deblurring, this section elaborates our \ourdataset~dataset, which is specially designed for human-aware motion deblurring.

\begin{table}
	\centering
	%\vspace{-4mm}
	\resizebox{0.49\textwidth}{!}{
		\setlength\tabcolsep{2pt}
		\renewcommand\arraystretch{1.10}	
		\begin{tabular}{c||cc|cc|cc}
			\hline\thickhline
    %\multirow{1}*{} &video                   &\!\!frames\!\!
    \rowcolor{mygray}
			&\multicolumn{2}{c|}{Quantity of People}
			&\multicolumn{2}{c|}{Depth of Object}
			&\multicolumn{2}{c}{Dataset Splits}\\
			\cline{2-7}
    \rowcolor{mygray}
			&&&Long-Shot& Close-Ups &&\\
    \rowcolor{mygray}	 \multirow{-3}*{\ourdataset}&\multirow{-2}*{~~Scattered~~}&\multirow{-2}*{Crowded}&(\ourdataset~I)&(\ourdataset~II)  &\multirow{-2}*{Train}&\multirow{-2}*{Test}\\
			\hline
            \hline
			\#~Images&4,202&4,220&1,304&7,118 &6,397 &2,025\\
            \#~FG Human& 23,474&42,310&14,169&51,615&44,318&21,466\\
			\hline
		\end{tabular}
	}
\vspace{-5pt}
\caption{\small Statistics of the proposed \ourdataset~dataset (see \S\ref{sec:dataset}).}
	\label{table:Statistics}
\vspace{-16pt}	
\end{table}

\noindent\textbf{Data Collection:}~Following the non-uniform blur generation methodology proposed in~\cite{Su_2017_CVPR,wieschollek2017learning}, we capture videos at 240fps with a GoPro Hero camera. Frames from these high-fps videos are then integrated to produce plausible motion blurs.
More specifically, as our dataset is designed for the multi-blur attention problem, we focus on collecting videos with humans in a close-up view to help facilitate moving-object annotation.
To guarantee the diversity of the dataset, we select various real-world scenarios with different quantities of humans. Blurred images are then synthesized by averaging 11 sequential frames from a video to simulate the degradation process, and the central frame is kept as the sharp image.

We clean the initial collection by taking into account two aspects. First, to account for hardware limitations, overly quick movements are equivalent to skip frames, resulting in streak artifacts in the blurred images.
Second, not all images contain an explicit structure or human in the close-up, especially if massive flat areas or pure scenes are present.
Thus, we remove candidates with these drawbacks, finally reaching 8,422 sharp and blurry image pairs in total. Fig.~\ref{fig:sample} shows some sample images from our dataset.

\noindent\textbf{Data Annotation:}~Unlike conventional pixel-wise tasks (\eg, segmentation and parsing), which preserve clear and sharp object edges for labeling, for our motion deblurring dataset, the FG humans are typically subject to motion displacements, and thus cannot be annotated with precise boundaries. Therefore, we annotate the FG humans in our dataset using bounding boxes. To improve annotation efficiency, we first apply a state-of-the-art human detection model~\cite{pose:fang2017rmpe} to each sharp image in our dataset, which can provide roughly accurate human bounding boxes for most human objects. Then, we manually refine the inferior results and add annotations for undetected humans.
To adequately apply the multi-motion blending model, we also remove the bounding boxes of BG humans in distant scenes to emphasize the close-up humans in the FG.

\noindent\textbf{Dataset Statistics:} Our \ourdataset~dataset has 8,422 sharp and blurry image pairs, extensively annotated with 65,784 human bounding boxes. The images are carefully selected from 31 high-fps videos and cover realistic outdoor scenes containing humans with various numbers, poses, and appearances at various distances (see Fig.~\ref{fig:sample}).

To describe this dataset in more depth, we present detailed attributes based on the quantity of humans in Table~\ref{table:Statistics}. The \textit{Scattered} subset consists of 4,202 scenes with a small number of FG humans. Analogously, the \textit{Crowded} set contains 4,220 images with large clusters of humans.

We subsequently organize the images into two categories, including long-shot (\ourdataset~I) and regular pedestrians (close-ups, \ourdataset~II), as shown in Table~\ref{table:Statistics}.
Evaluating each group can capture different aspects of the multi-motion blurring problem. For the \ourdataset~II dataset, as the FG human beings undergo more significant motions, it provides more emphasize on the challenges caused by FG actions.

\noindent\textbf{Dataset Splits:}~For evaluation purpose, the images are split into separate training and test sets (no overlap in source videos). Following random selection,
we arrive at a unique split containing 6,397 training and 2,025 test images. %Our dataset will be publicly released.

\vspace{-2pt}
\section{Proposed Algorithm}
\vspace{-3pt}
%In this section, we propose a human-aware attention model that exploits embedding features in terms of FG and BG to accelerate deblurring task.
%As discussed in Section~\ref{dataset}, a global motion blur generically causes by camera motion, in particular, close-ups which undergoing an apparent motion would yield a multiple degradation with respect to an object movement.
%%
%To address this problem, we introduce a discriminant mechanism to identify the property of respective blur in the blending motion model.
%%
%The architecture of the proposed human-aware deblurring network in Figure~\ref{fig:network}, we equip a neural attention network to infer a soft attention mask which enables a dynamic selection of FG/BG region from the blur images.
%
%In shark contrast, by leveraging a fully differentiable, human-aware attention module, the proposed model effectively disentangles the blur factors behind FG and BG, leading to a unified and end-to-end deblurring framework.
%
%Object motion which undergoing a camera motion as well as an object movement results in a multiple blur.
%%
%In particular, the motion usually yields large blur on the moving human in the FG.
\subsection{Attentive Motion Deblurring Model}\label{sec:overview}
\vspace{-3pt}

\begin{figure}[t]
%%tr = 0.006, ts = 0.008
  \centering
      \includegraphics[width=1 \linewidth]{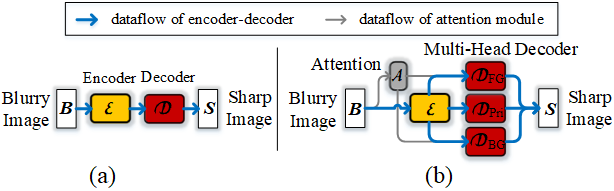}
\vspace{-15pt}
\caption{\small (a) A classical encoder-decoder based deblurring network. (b) Our proposed motion deblurring model, which is equipped with a human-aware attention module and a multi-head decoder. See \S\ref{sec:overview} for details.}
\label{fig:overall}
\vspace{-12pt}
\end{figure}

\noindent\textbf{Vanilla Encoder-Decoder based Deblurring Model:}~Our human-aware motion deblurring model is built upon a convolutional encoder-decoder network architecture (see Fig.~\ref{fig:overall}~(a)), which contains two parts, an encoder $\mathcal{E}$ and a decoder $\mathcal{D}$.
The encoder and decoder are comprised of a stack of convolutional and transposed convolutional layers, respectively, interleaved with non-linear point-wise nonlinearity (\eg, sigmoid).
The encoder aims to extract a new representation $\mathbf{H}\!\in\! \mathbb{R}^{w\times h\times c}$ from an input blurry image $\mathbf{B}\!\in\! \mathbb{R}^{W\times H\times 3}$, which is used by the decoder to predict the corresponding sharp image $\hat{\mathbf{S}}\!\in\! \mathbb{R}^{W\times H\times 3}$:
\vspace*{-6pt}
\begin{equation}
\begin{aligned}
\mathbf{H}&=\mathcal{E}(\mathbf{B};\mathbf{W}_{\mathcal{E}}),\\
\hat{\mathbf{S}}&=\mathcal{D}(\mathbf{H};\mathbf{W}_{\mathcal{D}}),
\end{aligned}
\label{eq:autoencoder}
\vspace*{-4pt}
\end{equation}
where $\mathbf{W}_{\mathcal{E}}$ and $\mathbf{W}_{\mathcal{D}}$ are a stack of learnable convolutional kernels for the encoder and decoder, respectively. The non-linear activation layers and bias term are omitted for convenience. %The embedded code $\mathbf{H}$ serves as the new representation of input blurry image, which is used for recovering the original image by the decoder.

%%%%%%%%%%%%%%%%%%% Figure 2%%%%%%%%%%%%%%%%%%%%%%

To explicitly encode FG human information into our model, we further introduce a supervised, human-aware attention mechanism into the encoder-decoder architecture (see Fig.~\ref{fig:overall}~\!(b)). Before delving into the details of our model, we first elaborate the proposed attention module.

\noindent\textbf{Human-Aware Attention Model:}~Here, we first provide a general formulation of differentiable neural attentions. Let $\mathbf{x}\!\in\!\mathbb{R}^{K\times C\!}$ be an input tensor, $\mathbf{z}\!\in\! \mathbb{R}^{k\times c\!}$ a feature obtained from $\mathbf{x}$, $\mathbf{a}\!\in\! [0,1]^{k\!}$ a soft attention vector, $\mathbf{g}\!\in\! \mathbb{R}^{k\times c\!}$ an attention-enhanced feature and $\mathcal{A}\!: \!\mathbb{R}^{K\times C\!}\!\rightarrow\! \mathbb{R}^{k\!}$ an attention network that learns to map $\mathbf{x}$ to a significance vector $\mathbf{y} \!\in\! \mathbb{R}^{k\!}$. The neural attention is implemented as:
  \vspace*{-4pt}
\begin{equation}\small
\begin{aligned}
\mathbf{a} &= \sigma(\mathbf{y}) = \sigma(\mathcal{A}(\mathbf{x})),\\
\mathbf{g} &= [\mathbf{a}\odot\mathbf{z}_1, \mathbf{a}\odot\mathbf{z}_2, \dots,\mathbf{a}\odot\mathbf{z}_c],
\end{aligned}
\label{eq:generalattention}
\vspace*{-6pt}
\end{equation}
where $\sigma$ indicates an activation function that maps the significance value into $[0,1]$, $\mathbf{z}_i\!\in\!\mathbb{R}^{k\!}$ indicates the feature in the $i$-th channel of $\mathbf{z}$, and `$\odot$' is an element-wise multiplication. The most popular strategy is to apply a softmax operation over $\mathbf{y}$ and learn (\ref{eq:generalattention}) in an \textit{implicit} manner.

In our approach, as we focus on image reconstruction, we extend the implicit attention above into a spatial domain. Similar to (\ref{eq:generalattention}), our human-aware attention network $\mathcal{A}\!: \!\mathbb{R}^{W\times H\times3}\!\rightarrow\! \mathbb{R}^{w\times h}$ learns to map the input blurry image $\mathbf{B}\!\in\! \mathbb{R}^{W\times H\times 3}$ to an importance map $\mathbf{Y} \!=\! \mathcal{A}(\mathbf{B})\!\in\! \mathbb{R}^{w\times h}$. An attention map $\mathbf{A}\!\in\! [0,1]^{w\times h}$ can be computed:
  \vspace*{-3pt}
\begin{equation}\small
\begin{aligned}
\mathbf{A} &= \sigma(\mathbf{Y}) = \text{sigmoid}(\mathcal{A}(\mathbf{B})).
\end{aligned}
\label{eq:ourattention}
\vspace*{-4pt}
\end{equation}
Since we have annotations for humans, which provide the groundtruth for the attention, we relax the sum-to-one constraint of softmax and instead use a \textit{sigmoid} activation function, \ie, only constrain the attention response values ranging from 0 to 1:
$\mathbf{A}_{i,j} = 1/(1+\exp(-\mathbf{Y}_{i,j}))$.

Then, we add supervision from human annotation over the attention map $\mathbf{A}$, \ie, we \textit{explicitly} train the attention network $\mathcal{A}$ by minimizing the following pixel-wise $\ell_2$ loss:
\vspace*{-5pt}
\begin{equation}\small
\mathcal{L}_{A} = \|\mathbf{G}-\mathbf{A}\|_2^2,
\label{eq:attentionloss}
%\vspace*{-4pt}
\end{equation}
where $\mathbf{G}\!\in\! \{0,1\}^{w\times h\!}$ is the binary FG human mask (see the small images in Fig.~\ref{fig:sample}). This way,
%our neural attention network devotes to synthesize a soft attention map to divide the FG and BG region.
the attention $\mathbf{A}$ encodes FG human information in a fully differentiable and supervised manner, which can be viewed as a soft FG mask.

%the attention network is trained for generating a mask which focuses on a sub-domain of the specific blur.

\noindent\textbf{Attention-Enhanced Encoder-Features:}~Then, to obtain an FG human-aware attention-enhanced feature $\mathbf{H}_{\text{FG}}\!\in\! \mathbb{R}^{w\times h\times c}$, we have:
  \vspace*{-4pt}
\begin{equation}\small
\begin{aligned}
\mathbf{H}_{\text{FG}} = [\mathbf{A}\odot\mathbf{H}_1,  \mathbf{A}\odot\mathbf{H}_2, \dots, \mathbf{A}\odot\mathbf{H}_c].
\end{aligned}
\label{eq:fgh}
\vspace*{-6pt}
\end{equation}
Similarly, we can obtain a soft BG mask through $(1\!-\!\mathbf{A})\!\in\! [0,1]^{w\!\times\! h\!}$, and further obtain a BG-aware attention-enhanced feature $\mathbf{H}_{\text{BG}\!}\!\in\! \mathbb{R}^{w\!\times\! h\!\times\! c}$:
  \vspace*{-4pt}
\begin{equation}\small
\begin{aligned}
\mathbf{H}_{\text{BG}} = [(1\!-\!\mathbf{A})\odot\mathbf{H}_1,  (1\!-\!\mathbf{A})\odot\mathbf{H}_2, \dots, (1\!-\!\mathbf{A})\odot\mathbf{H}_c].
\end{aligned}
%\HL{\mathbf{H}_{\text{BG}} = \mathbf{H} - \mathbf{H}_{\text{FG}}.}
\label{eq:bgh}
\vspace*{-6pt}
\end{equation}
In this way, FG human and BG information are encoded into the attention-enhanced features, $\mathbf{H}_{\text{FG}\!}$ and $\mathbf{H}_{\text{BG}}$, while the overall image information is stored in $\mathbf{H}$. %We treat the attention map as a soft mask to extract the feature of the FG/BG.

\begin{figure*}[t]
	\centering
	\footnotesize
	\includegraphics[width=0.99\textwidth]{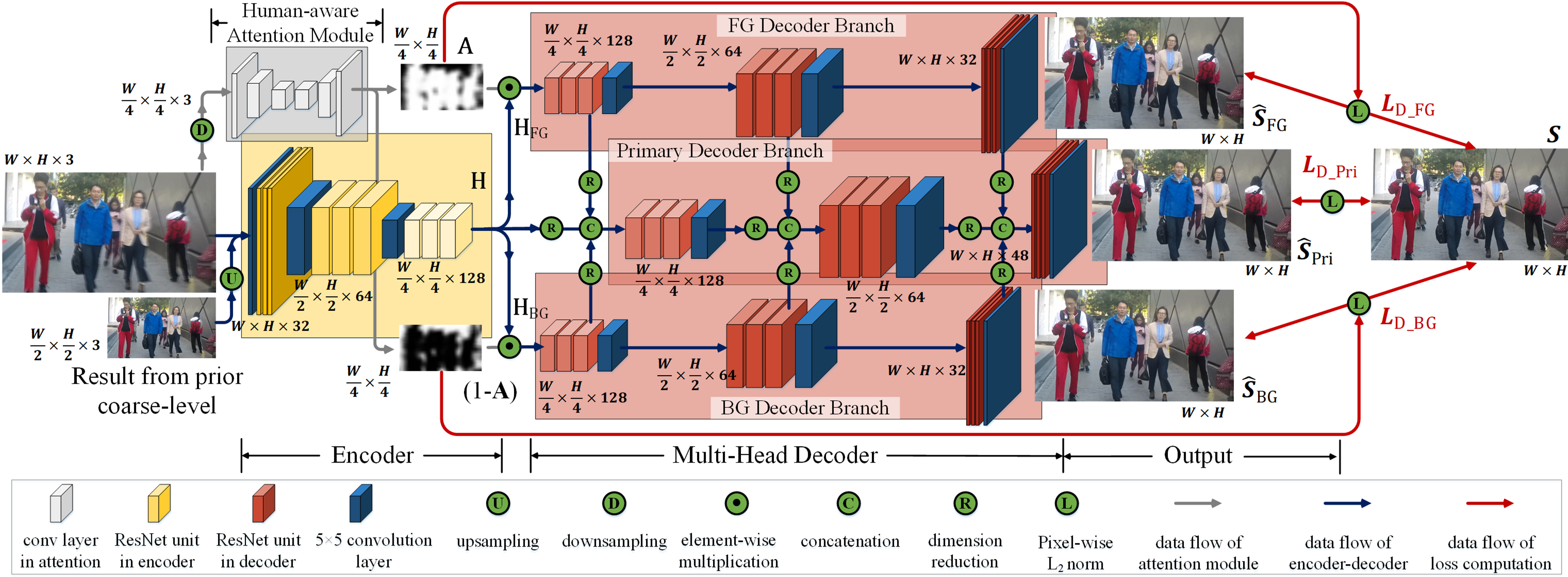}
\vspace*{-6pt}
	\caption{\small Overview of the proposed human-aware attentive deblurring network (single scale).}
\vspace*{-12pt}
	\label{fig:network}
\end{figure*}

\noindent\textbf{Multi-Head Decoder:}~With the original image feature $\mathbf{H}$ and enhanced features $\mathbf{H}_{\text{FG}}$ and $\mathbf{H}_{\text{BG}}$, we propose a multi-head decoder. As shown in Fig.~\ref{fig:overall}~\!(b), it is comprised of three branches: a primary one, an FG one and a BG one. Each branch takes the corresponding encoder features as inputs, and performs deblurring over the corresponding regions:
\vspace*{-2pt}
\begin{equation}\small
\begin{aligned}
\hat{\mathbf{S}}_{\text{FG}}&=\mathcal{D}_{\text{FG}}(\mathbf{H}_{\text{FG}}),\\
\hat{\mathbf{S}}_{\text{BG}}&=\mathcal{D}_{\text{BG}}(\mathbf{H}_{\text{BG}}),\\
\hat{\mathbf{S}}&=\mathcal{D}_{\text{Pri}}(\mathbf{H}).
\end{aligned}
\label{eq:mhd}
\vspace*{-1pt}
\end{equation}
For brevity, the corresponding learnable weights are omitted. The three decoder branches have similar network architectures (but without weight sharing). The critical role of this multi-head decoder module is preserving domain-specific features via individual FG/BG decoder branches.

To further encourage the FG decoder $\mathcal{D}_{\text{FG}}$ and BG decoder $\mathcal{D}_{\text{BG}}$ to focus on their corresponding regions, their deblurring loss functions are designed as:
\vspace*{-2pt}
\begin{equation}\small
\begin{aligned}
\mathcal{L}_{\text{D\_FG}}&=\mathbf{G}\odot\|\mathbf{S}-\hat{\mathbf{S}}_{\text{FG}}\|_2^2,\\
\mathcal{L}_{\text{D\_BG}}&=(1-\mathbf{G})\odot\|\mathbf{S}-\hat{\mathbf{S}}_{\text{BG}}\|_2^2.\\
\end{aligned}
\label{eq:branchloss}
\vspace*{-1pt}
\end{equation}
Take $\mathcal{D}_{\text{FG}}$ as an example.
By multiplying the squared error $\|\cdot\|$ with the binary FG human mask $\mathbf{G}$, the errors in the BG regions cannot be propagated back. This enables $\mathcal{D}_{\text{FG}}$ to handle FG blurs with more specific knowledge. Similarly, the employment of $(1\!-\!\mathbf{G})$ enables $\mathcal{D}_{\text{BG}}$ to concentrate more on deblurring of the background regions.

$\mathcal{D}_{\text{FG}}$ and $\mathcal{D}_{\text{BG}}$ capture domain-specific deblurring information, while the primary decoder $\mathcal{D}_{\text{Pri}}$ accounts for global information.
To make use of different deblurring information from different decoders in an ensemble manner, our idea is to use the specific knowledge from the $\mathcal{D}_{\text{FG}}$ and $\mathcal{D}_{\text{BG}}$ branches to support $\mathcal{D}_{\text{Pri}}$.
Instead of simply fusing their deblurring outputs (\ie, $\hat{\mathbf{S}}_{\text{FG}}, \hat{\mathbf{S}}_{\text{BG}}$,  and $\hat{\mathbf{S}}$) in a shallow manner, which might easily produce artifacts and inferior results, we use a deep knowledge-fusion strategy, \ie, inject multiple  intermediate features of $\mathcal{D}_{\text{FG}}$ and $\mathcal{D}_{\text{BG}}$ into $\mathcal{D}_{\text{Pri}}$.
More specifically, each decoder has a total of $L$ transposed convolutional blocks. Let us denote the features of the $l$-th block of $\mathcal{D}_{\text{FG}}$ ($\mathcal{D}_{\text{BG}}$) as $\mathbf{D}^l_{\text{FG}}$ ($\mathbf{D}^l_{\text{BG}}$) $\!\in\!\mathbb{R}^{w^l\times h^l\times c^l}$, where $l\!\in\!\{0,...,L\}$, the corresponding $l$-th layer feature of $\mathcal{D}_{\text{Pri}}$, can be recursively defined as:
\vspace*{-2pt}
\begin{equation}\small
\begin{aligned}
\mathbf{D}^l_{\text{Pri}} = \mathcal{D}_{\text{Pri}}^l\big(\langle\mathbf{D}^{l-1}_{\text{FG}}, \mathbf{D}^{l-1}_{\text{BG}}, \mathbf{D}^{l-1}_{\text{Pri}}\rangle\big),
\end{aligned}
\label{eq:primarybranch}
\vspace*{-1pt}
\end{equation}
where $\mathbf{D}^{0}_{\text{FG}}\!=\!\mathbf{H}_{\text{FG}}, \mathbf{D}^{0}_{\text{BG}}\!=\!\mathbf{H}_{\text{BG}}, \mathbf{D}^{0}_{\text{Pri}}\!=\!\mathbf{H}$, and $\langle\cdot\rangle$ indicates concatenation. For the final $L$-th layer of $\mathcal{D}_{\text{Pri}}$, we have:
\vspace*{-2pt}
\begin{equation}\small
\begin{aligned}
\hat{\mathbf{S}} = \mathbf{D}^L_{\text{Pri}}.
\end{aligned}
\label{eq:primaryS}
\vspace*{-1pt}
\end{equation}
As the primary decoder $\mathcal{D}_{\text{Pri}}$ comprehensively embeds domain-specific as well as global deblurring information, its loss function is designed over the whole image domain:
\vspace*{-2pt}
\begin{equation}\small
\begin{aligned}
\mathcal{L}_{\text{D\_Pri}}&=\|\mathbf{S}-\hat{\mathbf{S}}_{\text{Pri}}\|_2^2.
\end{aligned}
\label{eq:priloss}
\vspace*{-0pt}
\end{equation}
%Next, we will provide more details for our model.

\subsection{Detailed Network Architecture}\label{sec:architecture}
\vspace{-2pt}
Fig.~\ref{fig:network} illustrates the overall architecture of the proposed model. More details about the network architecture are provided in the supplementary material.

\noindent\textbf{Human-Aware Attention Module:}~Our human-aware attention module is created as a small network (see the gray blocks in Fig.~\ref{fig:network}). First, there are three convolutional layers, interleaved with $\times2$ max pooling and ReLU, which are stacked for effective image representation. Then, three transposed convolutional layers (each with a $\times2$ dilation rate and ReLU) are further adapted to enhance the image representation and spatial resolution. Finally, a
$1\! \times\! 1$ convolutional layer with sigmoid nonlinearity is added to produce an FG human prediction map $\mathbf{A}$ with the same size as the input image $\mathbf{B}$, using (\ref{eq:ourattention}).

\noindent\textbf{Encoder:}~The encoder module $\mathcal{E}$ consists of 9 residual units~\cite{he_residual} (the yellow blocks in Fig.~\ref{fig:network}). A $5\!\times\!5$ convolutional layer (the blue blocks in Fig.~\ref{fig:network}) is embedded between every three residual layers for dimensionality reduction. Then, we obtain a feature $\mathbf{H}$ and use the attention map $\mathbf{A}$ (a necessary downsampling operation is adopted) to obtain the enhanced features, $\mathbf{H}_{\text{FG}}$ and $\mathbf{H}_{\text{BG}}$, using (\ref{eq:fgh}) and (\ref{eq:bgh}), respectively.

\noindent\textbf{Multi-Head Decoder:}~With $\mathbf{H}$, $\mathbf{H}_{\text{FG}}$ and $\mathbf{H}_{\text{BG}}$, we further apply a multi-head decoder, which has three decoder branches, $\mathcal{D}$, $\mathcal{D}_{\text{FG}}$, and $\mathcal{D}_{\text{BG}}$, to reconstruct the input blurred image in their corresponding regions (see the red blocks in Fig.~\ref{fig:network}). Briefly, each of the branches has a structure symmetrical to the encoder network, \ie, comprising of nine transposed layers interleaved with dimensionality-reducing convolutional layers.
In addition, a shortcut connection between the encoder and each decoder module is embedded to compensate for generalization error.
Before fusing the enhanced features into the primary branch (see (\ref{eq:primaryS})), we first use a $1\! \times\! 1$ convolutional layer as a feature-compression layer. We then concatenate the enhanced features to compensate the final deblurring task, using (\ref{eq:primarybranch}). 	

\noindent\textbf{Multi-Scale Structure:}~A classical coarse-to-fine strategy is adopted, \ie, the single-scale model above is aggregated over three scales.
Here, weights are shared between scales to reduce the number of trainable parameters.
The multi-scale network goes through a continuous mechanism by integrating the degraded inputs with the previous result.
We extract features of each scale to enrich spatial information, and then extend the upper-scale representations by reusing the former collection.
For the first scale, the input blurry image is repeated to guarantee a feed-forward formulation.
We use the convolutional layer with a stride of 2 and $4\!\times\!4$ transposed convolutional layer to carry out downsampling and upsampling operations, respectively.

\vspace{-2pt}
\subsection{Implementation Details}
\vspace{-3pt}
\noindent\textbf{Training Settings:} We use the training sets of our \ourdataset~and the GoPro dataset. There are, in total, 10,742 training images with a size of $1280\!\times\!720$. The GoPro dataset is only used to train the BG decoder as it contains very few pedestrians. We crop a $256\!\times\!256$ patch for each image and use a batch size of 10 for each iteration. In addition, since BG takes a significant fraction in the training images, randomly cropping will cause an imbalance of training data for the BG and FG decoders. To alleviate this issue, the fractions of BG and pedestrians patch in each mini-batch are set to be harmonious. We use the Adam~\cite{adam} optimizer, with an initial learning rate of $1{e^{ - 4}}$.
The attention network is first pre-trained with 70,000 iterations for convergence. Then, the whole deblurring network is trained over 500 epochs.
% testing setting

\noindent\textbf{Reproducibility:}
Our model is implemented using Tensorflow~\cite{tensorflow}. All the experiments are done on a Titan X GPU. Our source code is released to provide full details of our training/testing processes and ensure reproducibility.

%\noindent\textbf{Testing and testing phases:}~Here we test the proposed framework on GoPro dataset and our \ourdataset, which contains a total of 3136 blurred frames with the size of $1280\!\times\!720$~(1111/2025 testing images in Gopro and \ourdataset~dataset, respectively). A experimental evidence would be provided to support the human-aware deblurring approach.

\vspace{-3pt}
\section{Experiments}
\vspace{-3pt}
In this section, we first perform an ablation study to evaluate the effect of each essential component of our model (\S\ref{sec:abation}).
Then, we provide comparison results with several state-of-the-art deblurring methods on the GoPro~\cite{nah2017deep} (\S\ref{sec:exgo}) and \ourdataset~(\S\ref{sec:exHIDE})~datasets.

\noindent\textbf{Evaluation Metrics:}~For quantitative evaluation, two standard metrics, Peak
Signal-to-Noise-Ratio (PSNR) and Structural Similarity Index
(SSIM), are adopted.

\noindent\textbf{Compared Methods:}~As we focus on motion deblurring, we include four state-of-the-art dynamic motion deblurring models~\cite{sun2015learning,Tao18,nah2017deep,Kupyn18} in our experiments. For a fair comparison, these models are also retrained using the training images of our \ourdataset~and the GoPro datasets.

\begin{table}[t]
	\centering
	\resizebox{0.48\textwidth}{!}{
		\setlength\tabcolsep{6pt}
		\renewcommand\arraystretch{1.0}
		\begin{tabular}{r||cc|cc|cc}
			\hline\thickhline
			\rowcolor{mygray}	&&&\multicolumn{4}{c}{\ourdataset}\\
			\cline{4-7}
			\rowcolor{mygray}
			&\multicolumn{2}{c|}{\multirow{-2}*{GoPro~\cite{nah2017deep}}}&\multicolumn{2}{c|}{\ourdataset~I}
			&\multicolumn{2}{c}{\ourdataset~II}\\
			\cline{2-7}
			\rowcolor{mygray}
			\multirow{-3}*{Methods~~~~}&PSNR &SSIM
			&PSNR &SSIM
			&PSNR  &SSIM \\
			\hline
			\hline
			\textbf{Ours} &\textbf{30.26}&\textbf{0.940}&\textbf{29.60}&\textbf{0.941}&\textbf{28.12}&\textbf{0.919} \\
			\hline
			\textit{w/o} attention&29.30 &0.929&28.40 &0.927&27.58&0.912\\
			FG branch&29.59 &0.931&28.59&0.928&27.55&0.909\\
			BG branch&29.78&0.934&28.89&0.931&27.68&0.911\\
			Single-scale & 29.85&0.934&28.47&0.930&27.81&0.916\\
			\hline
		\end{tabular}
	}
	\vspace{-5pt}
	\caption{\small Ablation study of our proposed human-aware deblurring model, evaluated on the GoPro~\cite{nah2017deep} and \ourdataset~datasets using PSNR and SSIM. See \S\ref{sec:abation} for details.}
	\vspace{-10pt}
	\label{tab:atten}
\end{table}%

\vspace{-2pt}
\subsection{Ablation Study}\label{sec:abation}
\begin{figure}[t]
	%%tr = 0.006, ts = 0.008
	\centering
	\includegraphics[width=1 \linewidth]{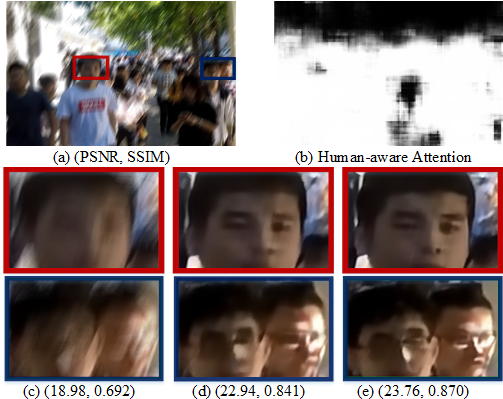}
	\vspace{-20pt}
	\caption{\small Demonstration of attention module for deblurring. (a) Blurred image. (b) Attention mask. (c) Blurred details. (d) Deblurred \textit{w/o} attention. (e) Deblurred \textit{w/} attention. See \S\ref{sec:abation}.}
	\label{fig:atten}
	\vspace{-6pt}
\end{figure}

\vspace{-3pt}
\noindent\textbf{Human-Aware Attention Module:} We first assess the impact of our human-aware attention module. We derive a baseline \textit{w/o attention} by retraining our model without the attention module. As demonstrated in Table~\ref{tab:atten}, the baseline \textit{w/o attention} clearly performs worse.
From Fig.~\ref{fig:atten}, we also observe that \textit{w/o attention} cannot restore an accurate profile (see (d)), while our full model gains better results (see (e)). This shows that the attention module enables the deblurring network to reconstruct an image with more facial features and accurate shape.
%
%Add the face-aware deblurring result for ablation study.

\begin{figure}[t]
	%%tr = 0.006, ts = 0.008
	\centering
	\includegraphics[width=1 \linewidth]{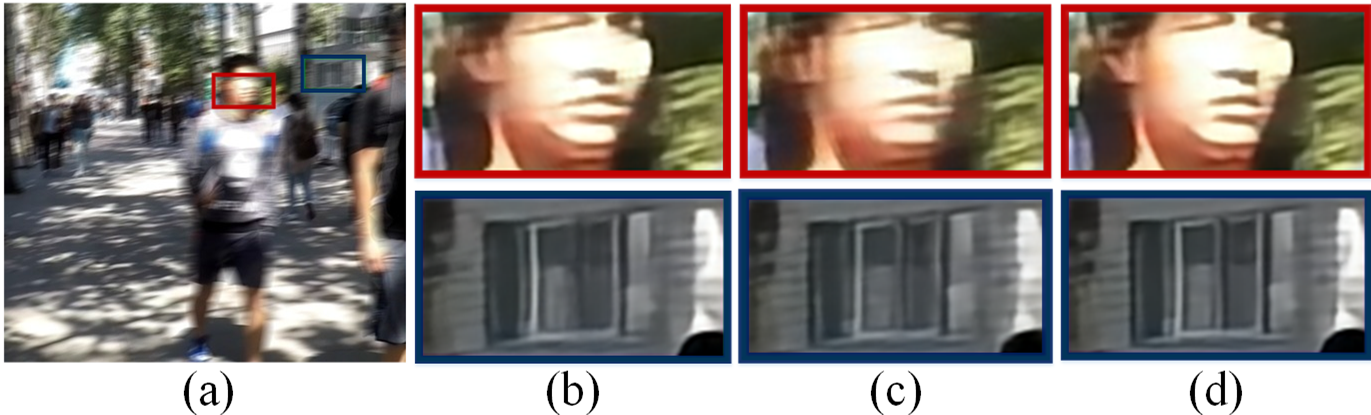}
	\vspace{-20pt}
	\caption{\small Effect of our multi-head structure.	(a) Blurred image. (b) Deblurred result of the FG decoder branch. (c) Deblurred result of the BG decoder branch. (d) Blending deblurred result. See \S\ref{sec:abation}.}
	\label{fig:branch}
	\vspace{-6pt}
\end{figure}

\noindent\textbf{Multi-Head Decoder:}~Next, to evaluate the effect of our multi-head decoder, we present a visual comparison between the deblurring results from different branches in Fig.~\ref{fig:branch}.
As shown in Fig.~\ref{fig:branch}(b) and (c), the FG deblurring branch and BG branch can handle blurring in their respective regions.
%
%While for the reversed region, as shown in Fig.~\ref{fig:branch}(b) deblur the BG region via FG reinforcing branch, it cannot be well restored.
We further compare them with the final blending result in Fig.~\ref{fig:branch}(d).
We find that, by inheriting complementary features from the FG and BG branches, the main branch can successfully restore the content in the full picture.
%
%A more detailed analysis of our multi-head deblurring architecture is described in the supplementary material.

%%%%%%%%%%%%%%%%%%% Figure 2%%%%%%%%%%%%%%%%%%%%%%
%%%%%%%%%%%%%%%%%%% Figure 2%%%%%%%%%%%%%%%%%%%%%%

\noindent\textbf{Multi-Scale Framework:}~As described in \S\ref{sec:architecture}, the proposed human-aware deblurring model works in a multi-scale fashion. To investigate the impact of such a design, we construct a \emph{single-scale} baseline and present the results in Table~\ref{tab:atten}.
As the proposed multi-scale model comes with better convergence, a faithfully reconstructed feature with respect to the latent image can be extracted, and the feed-forward mechanism is simultaneously applied to guide the network to generate better restoration results.
We provide more visual comparisons in the supplementary material.

\begin{table}[t!]
	\centering
	\resizebox{0.48\textwidth}{!}{
		\setlength\tabcolsep{4pt}
		\renewcommand\arraystretch{1.0}
		\begin{tabular}{r||cc|cc|cc}
			\hline\thickhline
			\rowcolor{mygray}	&&&\multicolumn{4}{c}{\ourdataset}\\
			\cline{4-7}
			\rowcolor{mygray}
			&\multicolumn{2}{c|}{\multirow{-2}*{GoPro~\cite{nah2017deep}}}&\multicolumn{2}{c|}{\ourdataset~I}
			&\multicolumn{2}{c}{\ourdataset~II}\\
			\cline{2-7}
			\rowcolor{mygray}
			\multirow{-3}*{Methods~~~~}&PSNR &SSIM
			&PSNR &SSIM
			&PSNR  &SSIM \\
			\hline
			\hline
			%Kim \etal~\cite{hyun2013dynamic}  &23.64 &0.8239\\
			Sun \etal~\cite{sun2015learning} &24.64 &0.843&23.21 &0.797&22.26&0.796\\
			Nah \etal~\cite{nah2017deep}  &28.49 &0.908 &27.43 &0.902&26.18&0.878\\
			Tao \etal~\cite{Tao18}  &30.26                                                                               &0.934&28.60 &0.928&27.35&0.907\\
			Kupyn \etal~\cite{Kupyn18} &26.93& 0.884&26.44& 0.890 & 25.37  &0.867\\
			\hline
			\textbf{Ours}  &\textbf{30.26}&\textbf{0.940}&\textbf{29.60}&\textbf{0.941}&\textbf{28.12}&\textbf{0.919} \\
			\hline
		\end{tabular}
	}
	\vspace{-5pt}
	\caption{\small Overall quantitative evaluation on the GoPro~\cite{nah2017deep} and \ourdataset~datasets using PSNR and SSIM. See \S\ref{sec:exgo} and \S\ref{sec:exHIDE}.}
	\vspace{-10pt}
	\label{tab:exgo}
\end{table}

\begin{figure*}[t]
	\centering
	\footnotesize
	\includegraphics[width=1\textwidth]{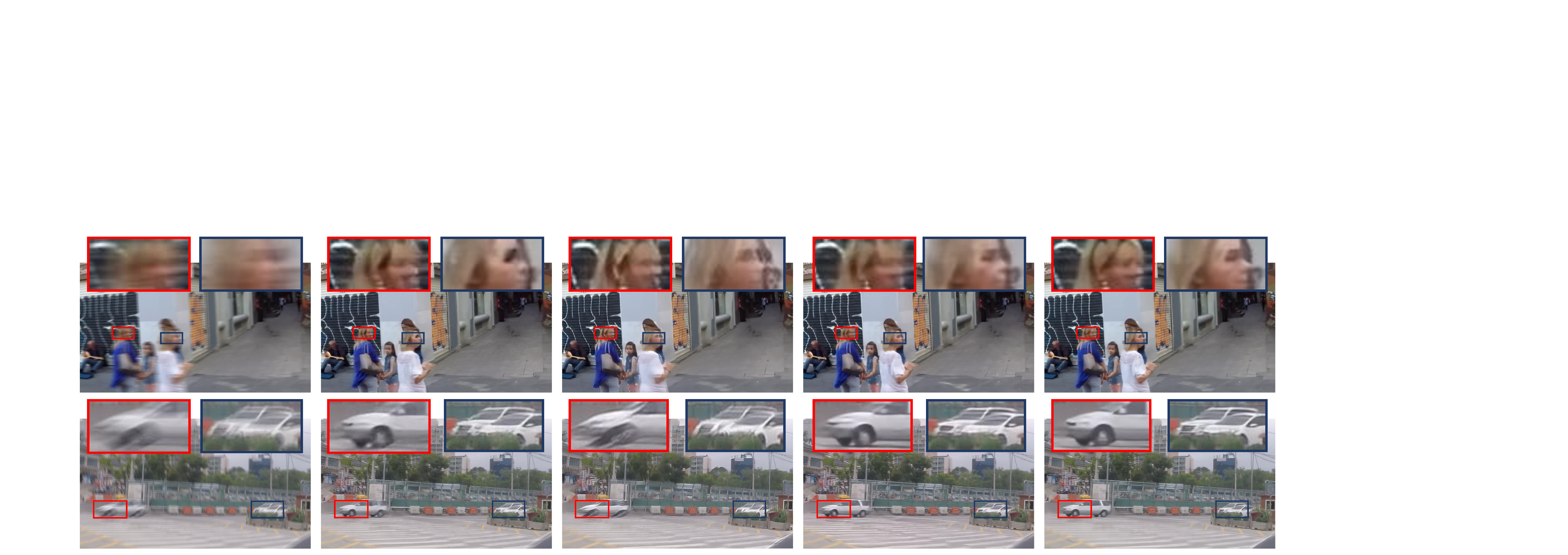}\vspace*{-2pt}
	\mbox{}\hfill (a) Blurred image \hfill\mbox{}
	\mbox{}\hfill (b) Nah \etal~\cite{nah2017deep}\hfill\mbox{}
	\mbox{}\hfill (c) Kupyn \etal~\cite{Kupyn18}\hfill\mbox{}
	\mbox{}\hfill (d) Tao \etal~\cite{Tao18}~~~~~\hfill\mbox{}
	\mbox{}\hfill (e) Ours~~~~~~~~~\hfill\mbox{}
	\vspace*{-6pt}
	\caption{\small Visual comparisons on the GoPro~\cite{nah2017deep} dataset (see \S\ref{sec:exgo}).}
	\vspace*{-8pt}
	\label{fig:exgo}
\end{figure*}

\begin{figure*}[t]
	\centering
	\footnotesize
	\includegraphics[width=1\textwidth]{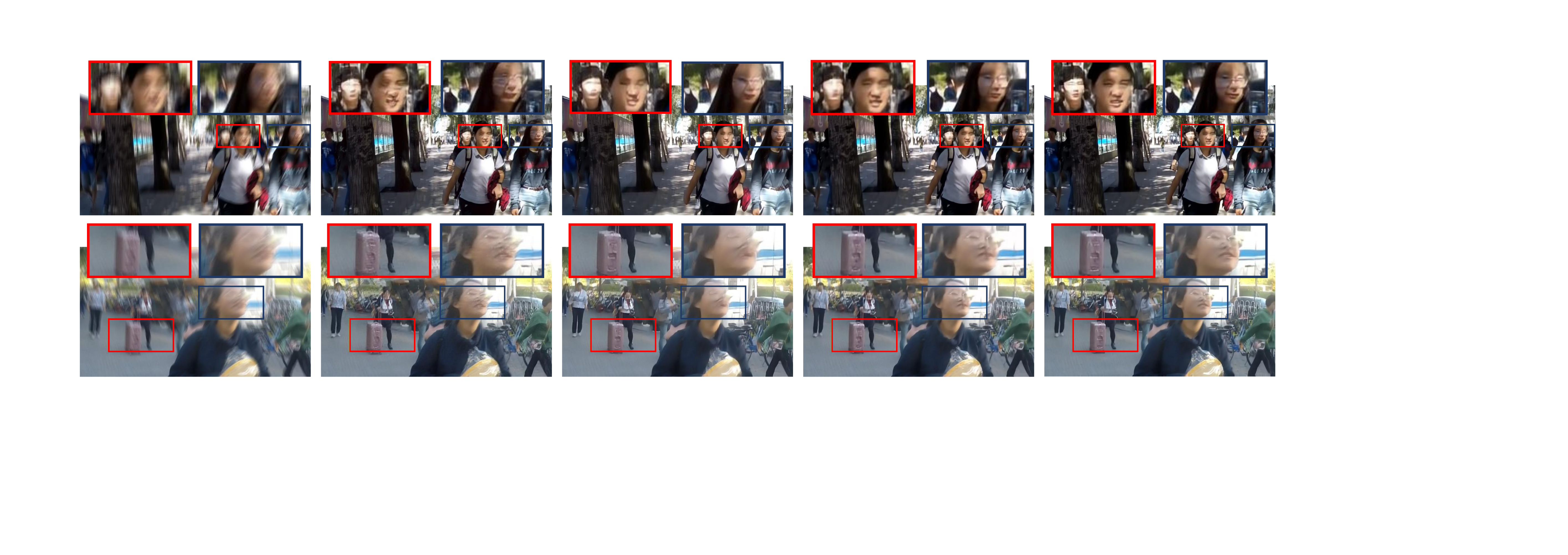}\vspace*{-2pt}
	\mbox{}\hfill (a) Blurred image \hfill\mbox{}
	\mbox{}\hfill (b) Nah \etal~\cite{nah2017deep}\hfill\mbox{}
	\mbox{}\hfill (c) Kupyn \etal~\cite{Kupyn18}\hfill\mbox{}
	\mbox{}\hfill (d) Tao \etal~\cite{Tao18}~~~~~\hfill\mbox{}
	\mbox{}\hfill (e) Ours~~~~~~~~~\hfill\mbox{}
	\vspace*{-6pt}
	\caption{\small Visual comparisons on our \ourdataset~dataset (see \S\ref{sec:exHIDE}).}
	\vspace*{-10pt}
	\label{fig:exHIDE}
\end{figure*}
\vspace{-2pt}
\subsection{Performance on the GoPro Dataset}\label{sec:exgo}
\vspace{-3pt}
We first evaluate the proposed model on the GoPro dataset~\cite{nah2017deep}, which contains 1,111 blurred images for testing.
Table~\ref{tab:exgo} shows a quantitative evaluation in terms of PSNR and SSIM, where our method shows promising results.
Furthermore, we provide a qualitative comparison in Fig.~\ref{fig:exgo}.
To verify the effectiveness of the proposed attentive deblurring framework, we first provide an example of a blurred image consisting of a moving human with an independent motion.
Our human-aware mechanism is able to perceive the specific movement and helps reconstruct a promising result with accurate face contours.
Moreover, as shown in the second row of Fig.~\ref{fig:exgo}, our method can improve deblurring in the full frame and perform well on a scaled scene thanks to the multi-head reconstruction module, which introduces a reinforcing strategy with two branches.
Overall, the proposed method applies a differentiable neural attention mechanism to a dynamic deblurring problem and achieves state-of-the-art performances.
\vspace{-3pt}
\subsection{Performance on the Proposed \ourdataset~Dataset}\label{sec:exHIDE}
\vspace{-3pt}
We note that the proposed method focuses mainly on the moving human deblurring problem, with multiple blurs caused by camera motion and/or human movement.
We further evaluate our approach on the \ourdataset~testing set.
In Fig.~\ref{fig:exHIDE}, we show visual comparisons, which relate to a specific human movement.
Due to degradation by complicated motion factors, the FG human undergoes a serious blur, and thus might not be accurately restored, \eg, with precise facial features and unambiguous outlines.
In contrast, the proposed human-aware attentive deblurring method exploits a multi-branch model to disentangle the FG human and BG. By fusing the reinforced features with the main branch, the network better restores the images with multiple blurs.

We also provide the associated qualitative comparison in Table~\ref{tab:exgo}.
The GoPro and \ourdataset~I datasets are mainly comprised of long-shot images, and hence involve only weak independent human movement.
By contrast, \ourdataset~II focuses on FG humans and provides a more comprehensive illustration for the moving human deblurring problem, for which our algorithm clearly outperforms previous state-of-the-arts.
More deblurring results are available in the supplementary material.

\vspace{-3pt}
\section{Conclusion}
\vspace{-3pt}
In this paper, we study the problem of human-aware motion deblurring. We first create a novel large-scale dataset dedicated to this problem, which is used in our study and expected to facilitate future research on related topics as well.
In addition, To handle multi-motion blur caused by camera motion and human movement, we propose a human-aware convolutional neural network for dynamic scene deblurring.
We integrate a multi-branch deblurring model with a supervised attention mechanism to reinforce the foreground humans and background, selectively.
By blending the different domain information, we restore the blurred images with more semantic details.
Experimental results show that the proposed approach performs favorably in comparison with state-of-the-art deblurring algorithms.

{\small
\bibliographystyle{ieee_fullname}
\bibliography{egbib}
}

\end{document}